\pdfoutput=1

\documentclass[11pt]{article}

\usepackage[]{acl}

\usepackage{times}
\usepackage{latexsym}
\usepackage{tabularx}
\usepackage{supertabular}
\usepackage{multirow}
\usepackage{algorithm}
\usepackage{algorithmic}
\usepackage{amssymb}
\usepackage{amsmath}
\usepackage{listings}
\usepackage{booktabs}
\usepackage{graphicx}
\usepackage{CJKutf8}
\usepackage{subfigure}

\usepackage[T1]{fontenc}

\usepackage[utf8]{inputenc}

\usepackage{microtype}

\usepackage{inconsolata}

\title{FPT: Feature Prompt Tuning for Few-shot Readability Assessment}

\author{Ziyang Wang$^{1,2}$,
        Sanwoo Lee$^{1,3}$, 
        Hsiu-Yuan Huang$^{1,3}$
        Yunfang Wu$^{1,3}$\thanks{~~Corresponding author.} \\
    $^{1}$National Key Laboratory for Multimedia Information Processing, Peking University \\ 
    $^{2}$School of Software and Microelectronics, Peking University, Beijing, China \\
    $^{3}$School of Computer Science, Peking University, Beijing, China \\
    \texttt{\{wzy232303, huang.hsiuyuan\}@stu.pku.edu.cn},
    \texttt{\{sanwoo, wuyf\}@pku.edu.cn},
    }

\begin{document}
\maketitle
\begin{abstract}
Prompt-based methods have achieved promising results in most few-shot text classification tasks. 
However, for readability assessment tasks, traditional prompt methods lack crucial linguistic knowledge, which has already been proven to be essential.
Moreover, previous studies on utilizing linguistic features have shown 
non-robust performance in few-shot settings and may even impair 
model performance.
To address these issues, we propose a novel prompt-based tuning framework that incorporates rich linguistic knowledge, called \textbf{F}eature \textbf{P}rompt \textbf{T}uning (FPT). Specifically, we extract linguistic features from the text and embed them into trainable soft prompts. Further, we devise a new loss function to 
calibrate the similarity ranking order between categories. 
Experimental results demonstrate that our proposed method FTP
not only exhibits a significant performance improvement over the prior best prompt-based tuning approaches, 
but also surpasses the previous leading methods that incorporate linguistic features.  
Also, our proposed model significantly outperforms the large language model gpt-3.5-turbo-16k in most cases. Our proposed method establishes a new architecture for prompt tuning that sheds light on how linguistic features can be easily adapted to linguistic-related tasks.
\end{abstract}

\section{Introduction}

Readability assessment (RA) is the task of evaluating the reading difficulty of a given piece of text~\cite{vajjala-2022-trends}. It has wide applications, such as choosing appropriate reading materials for language teaching~\cite{10.1145/1008992.1009112}, supporting readers with learning disabilities~\cite{rello-etal-2012-graphical}, and ranking search results by their reading levels~\cite{10.1145/2124295.2124323}.
\begin{figure}[thbp]
    \centering
    \includegraphics[width=\columnwidth]{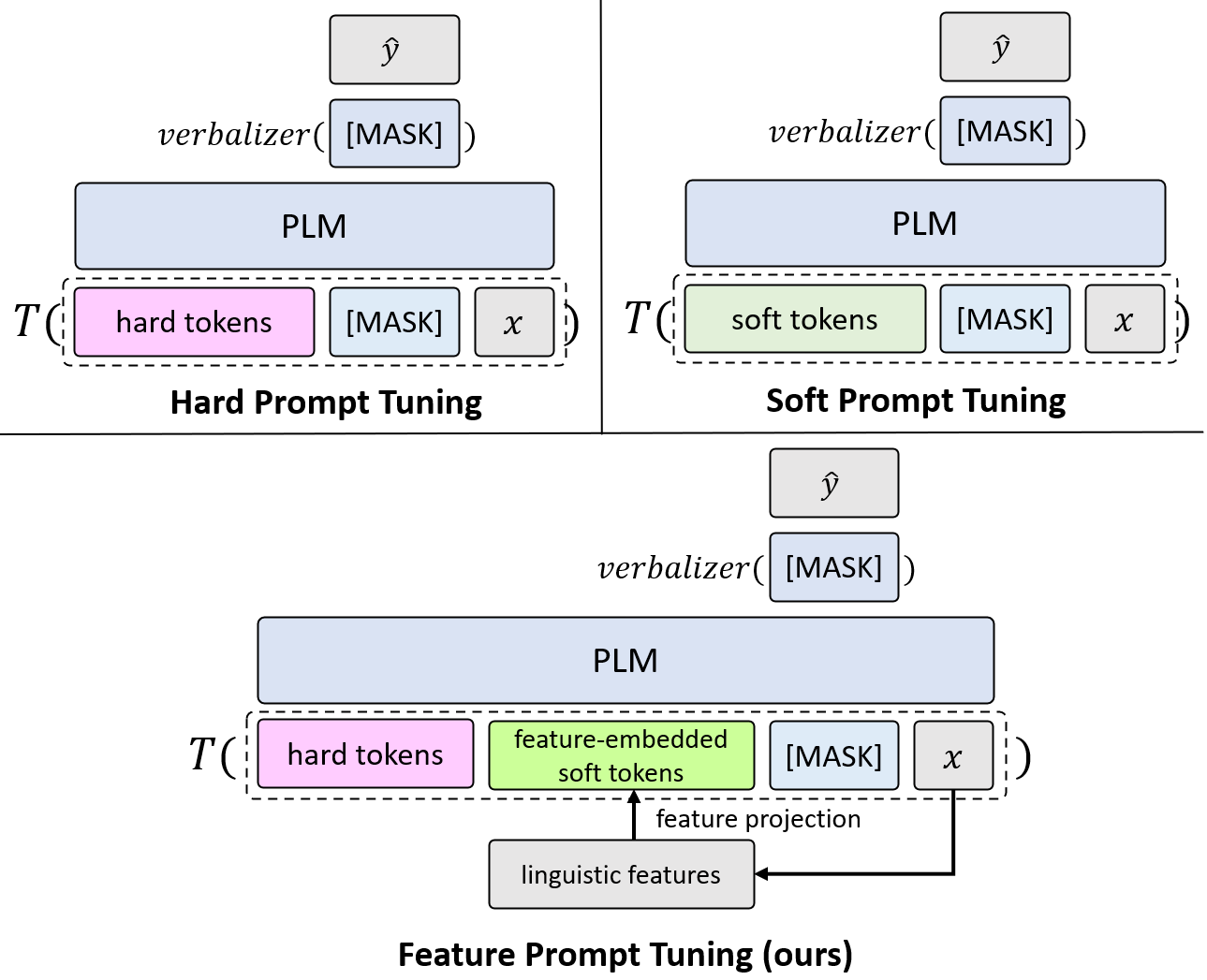}
    \caption{Comparison of previous prompt tuning frameworks and our proposed Feature Prompt Tuning (FPT). $T(\cdot)$ and $verbalizer(\cdot)$ denote the template and verbalizer function, respectively. FPT utilizes both hard and soft tokens which are projected from the linguistic features extracted from the input $x$.}
    \label{fig:Overview}
\end{figure}

Early works in RA mainly focused on designing handcrafted linguistic features such as word length (in characters/syllables), sentence length, and usage of different difficulty-level words.
In recent years, RA has been dominated by neural network-based architectures~\cite{DBLP:journals/corr/abs-2103-04083, azpiazu2019multiattentive}. The key challenge of these methods is to learn a better text representation that can capture deep semantic features. 
Current research has also explored different ways of combining linguistic features with pretrained language models (PLMs), achieving state-of-the-art results on numerous RA datasets~\cite{li-etal-2022-unified, lee2021pushing}. However, these studies have mainly focused on fine-tuning with a large amount of labelled data, while only a few studies have explored few-shot settings.

Prompt-based tuning, shown to be a powerful method for the classification task in the few-shot setting, makes full use of the information in PLMs by reformulating classification tasks as cloze questions. Different prompt-based tuning strategies are illustrated in Figure \ref{fig:Overview}. The hard prompt tuning 
applies a template with \text{[MASK]} token to the original input and maps the predicted label word to the corresponding class~\cite{HAN2022182,shin2020autoprompt}. 
The performance is sensitive to the quality of template, which introduces time-consuming and labor-intensive prompt design and optimization.
To address this problem, 
researchers propose soft prompt strategies, where continuous embeddings of trainable tokens replace the hard template and are optimized by training~\cite{liu2021gpt, lester-etal-2021-power}.

Despite the success in a range of text classification tasks, existing prompt-based tuning methods still suffer from inferior performance in RA. This might be attributed to the lack of linguistic knowledge which has been demonstrated to play a crucial role in RA~\cite{vajjala-2022-trends, qiu-etal-2021-learning, li-etal-2022-unified}. 
Meanwhile, RA differs from general classification tasks in that there exists a notion of ranking order between classes. Our intuition behind the utilization of linguistic knowledge is that the learned representations of different levels should preserve the similarity relationship analogous to that of original linguistic features of different levels.


Motivated by the above insights, in this paper, we propose a novel prompt-based tuning method that incorporates rich linguistic knowledge, called \textbf{F}eature \textbf{P}rompt \textbf{T}uning (FPT), as shown in the bottom of Figure~\ref{fig:Overview}. Specifically, our methodology begins with extracting linguistic features from the text. These extracted features are subsequently embedded into feature prompts, functioning as trainable soft prompts. Contrary to the conventional prompt tuning frameworks, our model can explicitly benefit from linguistic knowledge. Furthermore, we devise  
a new loss function to calibrate the similarity relationships between the embedded features across different categories. Our approach is straightforward and effective,
offering wide applicability to other tasks where the importance of handcrafted features is emphasized.

To verify the effectiveness of our proposed methods, we conduct extensive experiments on three RA datasets, including one Chinese data~\cite{li-etal-2022-unified} and two English datasets, WeeBit ~\cite{vajjala2012improving} and Cambridge~\cite{xia2019text}.  
By incorporating linguistic knowledge, our proposed model FPT improves significantly over other prompt-based methods. For instance, in the 2-shot setting, FPT brings a relative performance gain of 43.9\% over the traditional soft prompt method on the Chinese dataset and 5.50\% on English Weebit.   
Moreover, compared to 
other feature fusion methods, 
FPT outperforms the previous best method Projecting Feature (PF)~\cite{li-etal-2022-unified} by 43.19\% on Chinese data and 11.55\% on English Weebit data.
Also, we experiment on the Large Language Model (LLM), demonstrating the superiority of our approach on RA. 
We will make our code public available~\footnote{\url{https://github.com/Wzy232303/FPT}}.
We summarize our contributions as follows:
\begin{itemize}
    \item We propose a novel prompt-based tuning 
    framework, Feature Prompt Tuning (FPT), which incorporates rich linguistic knowledge for RA.
    \item We design a new calibration loss to ensure the linguistic features retain their original similarity information during optimization.
    \item Our experimental results show that our method outperforms other prompt-based tuning methods and effectively leverages linguistic features, leading to better and more stable performance improvements than previous approaches.
\end{itemize}
\section{Related Works}
\subsection{Readability Assessment}
Early works have explored a wide range of linguistic features as measurements for readability. \citet{flesch1948new} performed regression over features such as average word length in syllable; \citet{schwarm2005reading} trained an SVM over features including LM perplexity and syntactic tree height;  \citet{pitler2008revisiting} illustrated that discourse relations can be a good predictor of readability. 

Recent works largely employ deep learning approaches for RA. Several deep architectures, including BERT \cite{devlin2018bert}, HAN \cite{yang2016hierarchical}, and multi-attentive RNN were applied to achieve strong performance without feature engineering \cite{martinc2021supervised, azpiazu2019multiattentive}. However, the performance of neural models tends to fluctuate a lot across different RA datasets \cite{deutsch2020linguistic}, suggesting that relying only on neural networks might not be a robust solution for RA. 
Meanwhile, later works have shown that a hybrid approach combining transformer-based encoders with linguistic features can achieve even higher performance \cite{lee2021pushing, lee-vajjala-2022-neural, li-etal-2022-unified}. 
~\citet{lee-lee-2023-prompt} applied a prompt-based learning based on seq2seq model such as T5 and BART, treating RA as a text-to-text generative task. Despite the novelty of their method, it was not included in our baselines since it is hard for this method to draw a meaningful comparison against our approach. In addition to the fundamental discrepancy in the task definition, their method focuses on optimizing hard prompts and combining multiple datasets during training, whereas our method focuses on incorporating linguistic knowledge without leveraging multiple datasets. 


\subsection{Prompt-based Tuning}
Fine-tuning PLMs have shown their prevalence in various NLP tasks. PLMs, such as BERT \cite{devlin2018bert}, GPT \cite{radford2018improving}, XLNet \cite{xia2019text}, RoBERTa \cite{liu2019roberta} and T5 \cite{raffel2020exploring}, have been proposed with varied self-supervised learning architectures. It has been demonstrated that larger models tend to perform better in many learning scenarios \cite{brown2020language}, which stimulated PLMs with billions of parameters to emerge. 

Fine-tuning large PLMs may be prohibitive, and there exist a significant gap between pretraining tasks and downstream tasks. Prompt tuning addresses this challenge by reformulating downstream tasks as a language modeling problem and optimizing the prompt.
Prompts are used to probe PLM's intrinsic knowledge to perform a task \cite{min2022rethinking}, and various techniques of prompting have been explored to aid PLM better: hard prompt \cite{shin2020autoprompt, schick2021exploiting}, soft prompt \cite{lester-etal-2021-power, li2021prefix}, verbalizer \cite{cui2022prototypical} and pretrained prompt tuning \cite{gu2021ppt}.

The effectiveness of prompt tuning has been validated in various NLP tasks, including sentiment analysis \cite{wu2022adversarial}, named entity recognition \cite{ma-etal-2022-template}, relation extraction \cite{chen2022relation} and semantic parsing \cite{schucher2021power}. However, the potential of prompt tuning is less explored in RA.
In this work, we focus on the effectiveness of linguistic features for modeling readability, and utilize linguistic features to guide prompt tuning.
\section{Background}
We model RA as a text classification task. Formally, a RA dataset can be denoted as $\mathcal{D} = \{\mathcal{X}, \mathcal{Y}\}$, where $\mathcal{X}$ is the 
text set and $\mathcal{Y}$ is the class set. 
Each instance $x \in \mathcal{X}$ consists of several tokens, $x = \{w_1, w_2,...,w_{|x|}\}$, and is annotated with a label $y \in \mathcal{Y}$, indicating the reading difficulty. 
\begin{figure*}[htbp]
    \centering
    \includegraphics[width=\textwidth]{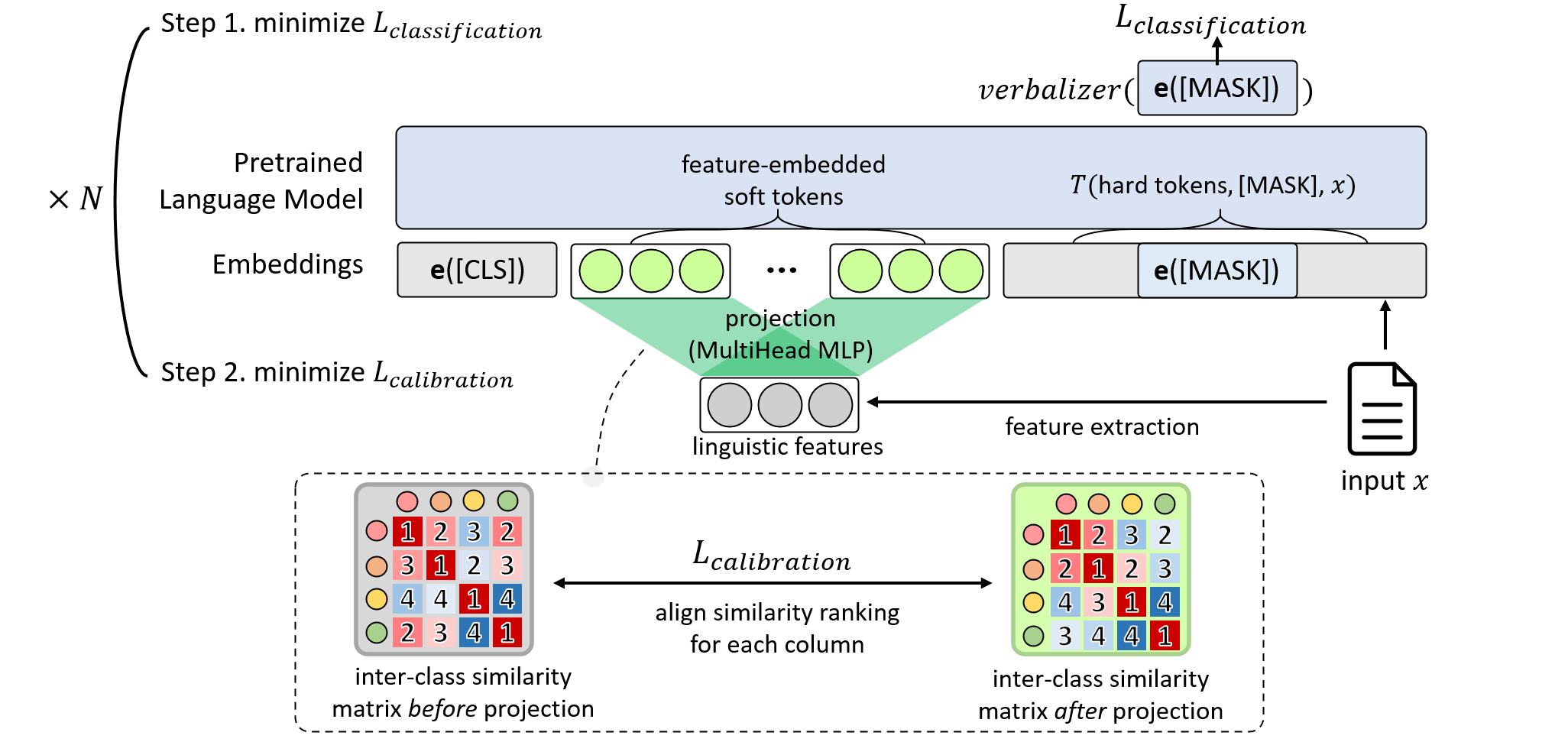}
    \caption{The architecture of the proposed Feature Prompt Tuning. Column-wise ranking orders of similarity matrices are denoted with numbers.}
    \label{fig:model}
\end{figure*}
\subsection{Fine-tuning PLMs for RA}
Given a PLM
$\mathcal{M}$ for RA, fine-tuning methods first convert 
a text $x = (w_1, w_2,...,w_{|x|})$ into an input sequence $([\text{CLS}], w_1, w_2,...,w_{|x|}, \text{[SEP]})$. The PLM encodes this sequence into 
the hidden vectors $h = (h_{[\text{CLS}]},h_1, h_2,...,h_{|x|},h_{\text{[SEP]}})$.

In the conventional fine-tuning, an additional classifier $FC$ is trained on top of the $[\text{CLS}]$ embedding $h_{[\text{CLS}]}$. This classifier 
produces a probability distribution over the class set $\mathcal{Y}$ through a softmax function, which can be formulated as:
$$ P(\cdot|x) = \text{Softmax}(FC(h_{[\text{CLS}]})),$$
The objective of fine-tuning is to minimize the cross-entropy loss between the predicted probability distribution $P(\cdot|x)$ and the ground-truth label $y$:
$$ \mathcal{L}_{classfication} = -\frac{1}{|\mathcal{X}|}\sum_{x \in \mathcal{X}} \log P(y|x). $$

\subsection{Prompt-based Tuning} \label{sec:PromptBasedTuning}
Prompt-based tuning aims to bridge the gap between pretraining tasks and downstream tasks, as illustrated in Figure \ref{fig:Overview}. 

\paragraph{Hard Prompt.} It typically consists of a template $T(\cdot)$, which transforms the input $x$ into a prompt input $x_{prompt}$, and a set of label words $V$ that are connected to the label space through a mapping function $\Phi: V \rightarrow \mathcal{Y}$, often referred to as the verbalizer. The 
prompt input contains at least one \text{[MASK]} token for the model to fill with label words. 

Taking an example in RA, $x_{prompt}$ could take the form of
$$
    x_{prompt}=T(x)=\text{"It is [MASK] to read: } x\text{"}.
$$
In this case, the input embedding sequence of $x_{prompt}$ is denoted as
$$
(e(\text{"It is"}), e(\text{[MASK]}), e(\text{"to read: "}), e(x)).
$$
\paragraph{Soft Prompt.} It replaces hard tokens in the template with trainable soft tokens $[h_1,...,h_l]$, yielding an input embedding sequence of 
$$(h_1, ... ,h_l, e(\text{[MASK]}), e(x)).$$

\paragraph{Hybrid Prompt.} 
It combines soft tokens with hard prompt tokens $T$
to form the input embedding sequence:
$$(h_1, ... ,h_l, e(T) ,e(\text{[MASK]}), e(x)).$$
By feeding the input embedding sequence of $x_{prompt}$  into $\mathcal{M}$, the probability distribution over the class set $\mathcal{Y}$ is modeled by:
$$ P_\mathcal{M}(y|x) = P_\mathcal{M}(\text{[MASK]} = \Phi(y)|x_{prompt}) $$
The learning objective of prompt-based tuning is to minimize the cross entropy loss:
$$ \mathcal{L}_{classification} = -\frac{1}{|\mathcal{X}|}\sum_{x \in \mathcal{X}} \log P_\mathcal{M}(y|x) $$

\section{Feature Prompt Tuning} 
In this section, we propose a novel method for RA with prompt-based tuning, named Feature Prompt Tuning (FPT). The architecture of our model is illustrated in Figure \ref{fig:model}.
Specifically, we extract linguistic features from the texts and embed them into soft prompts. Then, we employ a loss function to calibrate the similarity relationship between embedded features of different classes. We adopt an alternating procedure to optimize the model with respect to the classification loss and calibration loss.

\subsection{Feature Prompt Construction}\label{sec:FeaturePromptConstruction}

\paragraph{Feature Extraction}
Our approach for extracting linguistic features from text is consistent with previous works~\cite{li-etal-2022-unified, lee2021pushing}. For English texts, the linguistic features are extracted using the \textit{lingfeat} toolkit~\cite{lee2021pushing}, which includes discourse, syntactic, lexical, and shallow features. In terms of Chinese linguistic features, we directly utilize the \textit{zhfeat} toolkit~\cite{li-etal-2022-unified} to extract character, word, sentence, and paragraph-level features. Specific details are provided in Appendix \ref{sec:appendixFeatures}. 
For an input text $x$, we denote the extracted features as $f_x$, which is a $\alpha$-dimensional vector with $\alpha$ representing the number of extracted features. 

\paragraph{Feature Embedding}
To incorporate linguistic knowledge into prompt-based tuning, we transform linguistic feature $f_{x}$ into $l$ distinct vectors $\{v_1,...,v_l\}$ which function as the embeddings of soft tokens, as follows:
$$
\{v_1,...,v_l\} = \text{MultiHeadMLP}(f_x).
$$
Here, $\text{MultiHeadMLP}$ is a multi-head MLP with $l$ output heads. Each head consists of a series of fully connected layers followed by non-linear activation functions.


The purpose of using a multi-head MLP is to allow the model to map $f_{x}$ into separate vector spaces and learn 
multiple aspects of the linguistic features. This enables the model to better capture the relationships between different features and their contribution to RA.

Ultimately, we formulate the input embedding sequence of $x_{prompt}$ as follows:
$$
 (v_1, ... ,v_l, e(T) ,e(\text{[MASK]}), e(x)).
$$
This input sequence is passed through the PLM $\mathcal{M}$ to calculate $\mathcal{L}_{classification}$ loss as described in Section \ref{sec:PromptBasedTuning}.

\subsection{Inter-class Similarity Calibration}\label{sec:SimilarityCalibration}


We denote $\mathcal{F} = \{F_{c_1}, \cdots, F_{c_n}\}$ as the collection of linguistic features for the dataset $\mathcal{D}$, which consists of $n$ classes. Here, $F_{c_i} = \{f_{x_{i1}}, \cdots, f_{x_{ik}}\}$ signifies the extracted features of $k$ samples which belong to $i$-th class. We apply average pooling to the feature embeddings of each sample in $\mathcal{F}$, resulting in a set of embedded linguistic features, denoted as $\mathcal{F'} = \{F'_{c_1}, \cdots, F'_{c_n}\}$. To gauge the similarity between any two classes $F_{c_m}$ and $F_{c_n}$, we employ a pairwise approach based on cosine similarity, expressed as:
$$s_{mn} = \frac{1}{k^2}\sum_{i=1}^{k}\sum_{j=1}^{k}\cos(f_{x_{mi}}, f_{x_{nj}})$$

With the feature representation and similarity function in place, we can
define our calibration objective. The fundamental intuition is that the distribution of extracted linguistic features should be preserved as much as possible. Namely, 
if the similarity between $F_{c_m}$ and $F_{c_n}$ is relatively low, the similarity between $F'_{c_m}$ and $F'_{c_n}$ should also be proportionately low, and vice versa. 
Therefore, during the training process, we 
devise an objective function based on a list-wise ranking loss function ListMLE~\cite{10.1145/1390156.1390306}, 
to maintain this initial ranking relationship. 

More specifically, we compute the similarity between each pair of classes within $\mathcal{F}$ to generate the similarity matrix:
$$M = \left[ \begin{array}{cccc} s_{11} & s_{12} &\cdots & s_{1n} \\ s_{21} & s_{22} &\cdots & s_{2n} \\ \vdots & \vdots & \ddots & \vdots \\ s_{n1} & s_{n2} &\cdots & s_{nn} \end{array} \right]$$
Likewise, we can derive the similarity matrix $M'$ for $\mathcal{F'}$.

We then use $\Pi = \{ \pi_{1}, \pi_{2} , \cdots , \pi_{n} \}$ to denote the ranking order of the columns in $M$, where $\pi_{i}$ represents the ranking order of the $i$-th column. We obtain $\hat{M'}$ by rearranging the columns of $M'$ according to $\Pi$:
$$
\hat{M'} = \left[ \begin{array}{cccc} s'_{\pi_{11}} & s'_{\pi_{12}} &\cdots & s'_{\pi_{1n}} \\ s'_{\pi_{21}} & s'_{\pi_{22}} &\cdots & s'_{\pi_{2n}} \\ \vdots & \vdots & \ddots & \vdots \\ s'_{\pi_{n1}} & s'_{\pi_{n2}} &\cdots & s'_{\pi_{nn}} \end{array} \right]
$$

Finally, we aim to minimize the following loss function for similarity calibration:
$$
L_{calibration} = -\sum_{k=1}^{n}log\prod_{i=1}^{n}\frac{\exp(s'_{\pi_{ik}})}{\sum_{j=i}^{n}\exp(s'_{\pi_{jk}})}
$$

\subsection{Training Procedure}
\paragraph{Training Objectives}
Given the dataset $\mathcal{D}$ and the linguistic feature set
$\mathcal{F}$, we establish two training objectives. The primary objective is to minimize the classification loss $L_{classification}$, which is computed based on the difference between the predicted and actual class labels. The secondary objective is to calibrate the inter-class similarity of the mapped features by minimizing the loss function $L_{calibration}$ defined in Section \ref{sec:SimilarityCalibration}.

\paragraph{Alternating Training Procedure}
\begin{algorithm}[h]
\caption{Alternating Training Procedure for Feature Prompt Learning}
\label{alg:alternating_training}
\begin{algorithmic}[1]
\STATE Initialize model parameters $M$ and feature embeddings $f$
\FOR{each epoch}
    \STATE Shuffle dataset $D$
    \FOR{each batch $b$ in $D$}
        \STATE Compute $L_{classification}$ for $b$ using $M$ and $f$
        \STATE Update $M$ and $f$ by minimizing $L_{classification}$
    \ENDFOR
    \STATE Compute $L_{calibration}$ for $D$ using $f$
    \STATE Update $f$ by minimizing $L_{calibration}$
\ENDFOR
\end{algorithmic}
\end{algorithm}

For training both loss functions, we adopt an 
alternating training procedure, as encapsulated in Algorithm~\ref{alg:alternating_training}. This procedure iteratively updates the model parameters and feature embeddings by minimizing the classification loss $L_{classification}$ and the similarity calibration loss $L_{calibration}$, respectively.

In each epoch, the dataset $D$ is shuffled to ensure the model is not biased towards any particular ordering of the data. For each batch $b$ in $D$, the classification loss $L_{classification}$ is computed using the current model parameters $M$ and feature embeddings $f$. The model parameters $M$ and feature embeddings $f$ are then updated by minimizing this loss. Subsequently, the similarity calibration loss $L_{calibration}$ is computed using the updated feature embeddings $f$ for the epoch, and the feature embeddings are updated by minimizing this loss . This process is repeated for each epoch. The alternating training procedure ensures that the model learns to classify the data accurately while maintaining the inter-class similarity structure of the feature space.
\section{Experimental Setting}

\subsection{Datasets}
To evaluate the effectiveness of our proposed method, we conduct experiments on one Chinese dataset and two English datasets, following the same data split as ~\citet{li-etal-2022-unified}. The statistics of the datasets can be found in Table~\ref{table:dataset_statistics}.

\textbf{ChineseLR }~\cite{li-etal-2022-unified} is a Chinese dataset collected from textbooks of the middle and primary schools of more than ten publishers. Following the standards specified in the \textit{Chinese Curriculum Standards for Compulsory Education}, all texts are divided into five difficulty levels.

\textbf{WeeBit }~\cite{vajjala2012improving} is often considered as the benchmark data for English RA. It was initially created as an extension of the well-known Weekly Reader corpus. 

\textbf{Cambridge }~\cite{xia2019text} 
consists of reading passages from the five main suite Cambridge English Exams (KET, PET, FCE, CAE, CPE).


\subsection{Baselines 1: Prompt-based Methods}
For prompt-based methods, we compare with hard, soft, and hybrid prompts.
To avoid the influence of verbalizers on experimental results, we adopt a soft verbalizer~\cite{hambardzumyan-etal-2021-warp} that employs a linear layer classifier across all prompt-based methods. 

\textbf{Hard Prompt (HP)}: We implement 
four manually defined templates for prompt tuning and select a template with the best performance on the development set. As for FPT in Table~\ref{tab:promptresults}, we report the test set performance averaged over the four templates. This setting poses a challenge to FPT, as the averaged performance of FPT should outperform the best performance of HP to demonstrate its effectiveness. 
Details of the templates can be found in Appendix~\ref{sec:appendixTemplates}. 

\textbf{Soft Prompt (SP)}: 
It replaces manually defined prompts with trainable continuous prompts. We follow the same implementation as~\citet{lester-etal-2021-power} and use randomly sampled vocabulary to initialize the prompts. 

\textbf{Hybrid Prompt (HBP)}: It concatenates trainable continuous prompts to the wrapped input embeddings. We adopt the implementation from~\citet{gu-etal-2022-ppt}.

\textbf{P-tuning}: A hybrid prompt method, which replaces some tokens in manually designed prompts with soft prompts and only retains task-relevant anchor words. The soft prompts are embedded with a bidirectional LSTM and a MLP~\cite{liu2021gpt}.

\begin{table}[t]
\centering
\small
\resizebox{\linewidth}{!}{
\begin{tabular}{ccccccc}
\hline
\textbf{Dataset} & \multicolumn{2}{c}{\textbf{WeeBit}} & \multicolumn{2}{c}{\textbf{Cambridge}} & \multicolumn{2}{c}{\textbf{ChineseLR}} \\
\hline
\textbf{Level} & \#    & Avg Len        & \#           & Avg Len          & \# & \multicolumn{1}{c}{Avg Len} \\
\hline
1 & 625 & 152 & 60 & 141 & 814 & 266 \\
2 & 625 & 189 &  60 & 271 & 1063 & 679 \\
3 & 625 & 295 & 60 & 617 & 1104 & 1140 \\
4 & 625 & 242 &  60 & 763 & 762 & 2165 \\
5 & 625 & 347 &  60 & 751 & 417 & 3299 \\
All & 3125 & 245  & 300 & 509 & 4160 & 1255 \\
\hline
\end{tabular}
}
\caption{Statistics of RA datasets. \#: number of the passages. Avg Len: average tokens numbers per passage.}
\label{table:dataset_statistics}
\end{table}

\subsection{Baselines 2: Fusion Methods}
We also compare with the methods fusing linguistic features and PLMs from previous studies. 

\textbf{SVM}: 
Use the single numerical output of a neural model (BERT) as a feature itself, joined with linguistic features, and then fed them into SVM
~\cite{lee2021pushing,deutsch2020linguistic}.

\textbf{FT}: Standard fine-tuning method without linguistic features, where the hidden representation of [\text{CLS}] token is used for classification. This baseline validates whether the linguistic features indeed have a positive effect.

\textbf{Concatenation (Con)}: Fine-tune with linguistic features, in which the linguistic features are directly concatenated to the hidden representation of the [\text{CLS}] token ~\cite{DBLP:journals/corr/abs-2103-04083,qiu-etal-2021-learning}.

\textbf{PF}: 
Fuse linguistic features with hidden representations of [\text{CLS}] through projection filtering~\cite{li-etal-2022-unified}.

\subsection{Implementation Details}
Under the few-shot setting, we randomly sample $k=1, 2, 4, 8, 16$ instances in each class from the training and development set. For each $k$-shot experiment, we sample 4 different training and dev sets and repeat experiments on each training set for 4 times. We select the best model checkpoint based on the performance of the development set and evaluate the models on the entire test set. 
As for the evaluation metric, we use $accuracy$ in all experiments and take the mean values as the final results.

All our models and baselines are implemented with the PyTorch~\cite{paszke2019pytorch} framework and Huggingface transformers~\cite{wolf-etal-2020-transformers}. We use BERT~\cite{devlin2018bert} as our Pretrained Language Model (PLM) backbone. We use "bert-base-uncased" for English datasets and "bert-base-chinese" for the Chinese dataset.
During training, we employ the AdamW optimizer~\cite{loshchilov2018decoupled} with a weight decay of 0.01 and a warm-up ratio of 0.05. We tune the model with the batch size of 8 for 30 epochs, and the learning rate is 1e-5.  All experiments are conducted with four NVIDIA GeForce RTX 3090s.

\section{Results and Analysis}
\begin{table}[h!]
\centering
\small
\resizebox{\linewidth}{!}{
\begin{tabular}{clccc}
\toprule
\textbf{\textit{k}} & \textbf{Methods} & \textbf{ChineseLR} & \textbf{Weebit} & \textbf{Cambridge}  \\
\midrule
\multirow{5}{*}{1}
& HP & 29.49(5.21)  & 41.83(4.72) & 36.25(8.49)  \\
& SP  & 31.22(4.70)   & \underline{\textbf{46.61}}(3.63) &  41.73(8.45)  \\
& HBP  & \underline{33.51}(5.19)  & 44.46(5.02) &   \underline{42.04}(9.12)  \\
& P-tuning  & 33.36(4.12)  & 41.23(4.11) & 40.36(7.15)   \\
\cmidrule{2-5}
& FPT(ours) &  \textbf{39.63}(6.38) & 43.61(4.50) & \textbf{44.17}(7.12)  \\
\midrule
\multirow{5}{*}{2}
& HP & 28.38(8.14)  & 49.23(2.85) &  46.88(9.31)  \\
& SP  &  32.14(5.54)  & 52.22(4.35) & 49.13(8.38) \\
& HBP  & 33.38(7.02)   &  \underline{52.52}(2.66) & \underline{49.56}(7.12)  \\
& P-tuning  & \underline{35.12}(4.20)   &  50.71(3.87) &  48.97(8.47)  \\
\cmidrule{2-5}
& FPT(ours) &  \textbf{46.24}(5.62) & \textbf{55.10}(4.04) & \textbf{59.79}(10.2)  \\
\midrule
\multirow{5}{*}{4}
& HP & 36.56(5.18)  & 53.41(4.50)  & 48.75(8.49)  \\
& SP  &  38.78(2.83)  &  54.96(3.89) & 49.36(9.14)  \\
& HBP  &  \underline{39.81}(2.67)   & \underline{56.88}(3.52) & \underline{50.13}(8.77)   \\
& P-tuning  &  38.45(3.09)  & 54.35(3.21) &  48.85(9.64)  \\
\cmidrule{2-5}
& FPT(ours) & \textbf{48.93}(3.21)  & \textbf{57.70}(4.63) &\textbf{ 53.54}(7.21)  \\
\midrule
\multirow{5}{*}{8}
& HP &  41.21(4.83) & 61.31(3.13) & 55.42(6.86)  \\
& SP  &  42.72(2.82)  & 62.02(2.67) &  56.75(6.89) \\
& HBP  &  41.93(4.12)  & \underline{63.37}(2.02) &  \underline{57.34}(9.28)  \\
& P-tuning  &  \underline{42.81}(4.04)  & 61.81(3.28) & 56.90(7.23)   \\
\cmidrule{2-5}
& FPT(ours) & \textbf{52.66}(5.00)  & \textbf{64.92}(2.75) & \textbf{59.38}(6.58)  \\
\midrule
\multirow{5}{*}{16}
& HP & 47.35(3.69)  & 63.75(5.41) &  61.67(8.98) \\
& SP  &  \underline{47.44}(2.09)  & \underline{67.54}(4.56) &  63.77(7.43) \\
& HBP  &  47.08(3.11)  & 67.30(4.69) &  \underline{63.98}(7.34)  \\
& P-tuning  & 46.26(3.19)  & 65.52(3.84) & 62.03(9.62)   \\
\cmidrule{2-5}
& FPT(ours) & \textbf{55.25}(2.93)  & \textbf{68.19}(4.21) & \textbf{65.00}(4.25)  \\
\bottomrule
\end{tabular}}
\caption{Experimental results comparing with prompt-based methods.
We report the mean performance and the standard deviation in brackets. 
The best results are in bold, and the best results of previous prompt-based methods are underlined.} 
\label{tab:promptresults}
\end{table}

\begin{table}[h!]
\centering
\small
\resizebox{\linewidth}{!}{
\begin{tabular}{clccc}
\toprule
\textbf{\textit{k}} & \textbf{Methods} & \textbf{ChineseLR} & \textbf{Weebit} & \textbf{Cambridge}  \\
\midrule
\multirow{6}{*}{1}
& FT & 28.59(4.88)  &\underline{45.99}(2.94)  &  34.17(4.33) \\
& SVM  &  25.34(3.87)  & 44.82(3.14) &  \underline{35.31}(5.23) \\
& Con  & 28.53(4.68)   & 43.81(3.88) & 33.33(10.1)   \\
& PF & \underline{30.13}(3.99)   & 44.01(2.91) & 35.11(9.12)  \\
\cmidrule{2-5}
& FPT(ours) &\textbf{ 33.29}(4.80)  & \textbf{46.67}(3.50) & \textbf{43.96}(7.09)  \\
\midrule
\multirow{6}{*}{2}
& FT & 22.87(7.19)  & 48.79(3.49) & \underline{44.17}(10.4)  \\
& SVM  &  23.95(9.28)  & 49.55(3.78) &  43.99(11.0) \\
& Con  & 25.61(8.21)   & 49.29(2.88) &  41.67(8.16)  \\
& PF & \underline{26.12}(7.21)   & \underline{50.23}(2.81) &  41.52(7.34)   \\
\cmidrule{2-5}
& FPT(ours) & \textbf{37.40}(4.77)  & \textbf{56.03}(3.48) & \textbf{55.83}(6.72)  \\
\midrule
\multirow{6}{*}{4}
& FT & 36.64(5.37)  & 52.46(4.28) & 47.50(6.29)  \\
& SVM  & 37.11(6.88)  & 53.03(5.65) & 47.58(8.67)  \\
& Con  &  36.64(5.37)  & 52.46(4.28) & 47.50(6.29)  \\
& PF &\underline{37.13}(5.11)  & \underline{53.18}(2.99) & \underline{48.46}(4.79)   \\
\cmidrule{2-5}
& FPT(ours) & \textbf{44.88}(3.27)  & \textbf{56.17}(3.84) & \textbf{55.00}(4.86)  \\
\midrule
\multirow{6}{*}{8}
& FT & 40.45(2.91)  & \underline{61.11}(3.15) &  61.46(7.81) \\
& SVM  & 40.52(3.67)   & 60.98(5.78) & \underline{61.55}(9.10)  \\
& Con  &   41.65(2.98)  & 58.41(3.31) &  58.96(7.43)  \\
& PF & \underline{44.00}(2.86)   & 59.32(2.97)  &  55.62(10.9)  \\
\cmidrule{2-5}
& FPT(ours) & \textbf{47.60}(3.66)  & \textbf{62.40}(3.30) & \textbf{64.17}(5.95)  \\
\midrule
\multirow{6}{*}{16}
& FT & 45.73(4.11)  & \underline{65.93}(5.50) & 71.04(7.97)  \\
& SVM  & 46.85(3.72)  & 63.72(4.98)  & 71.22(8.15)  \\
& Con  & 48.33(3.99)   & 64.52(4.73) & \underline{\textbf{71.46}}(6.12)   \\
& PF &  \underline{48.66}(3.20)    & 65.08(4.60)  &  69.38(6.79)  \\
\cmidrule{2-5}
& FPT(ours) & \textbf{53.94}(3.16)  & \textbf{68.10}(3.25) & 69.17(7.77)  \\
\bottomrule
\end{tabular}}
\caption{Experimental results 
comparing with feature fusion methods. Con means Concatenation. For a fair comparison, here 
FPT
concatenates the embedded linguistic features to the embeddings of the input sequence (without hard prompt template) and outputs the classification logits over $[\text{CLS}]$ token embedding instead of $[\text{MASK}]$.}
\label{tab:finetuneresults}
\end{table}

\subsection{Comparison with Prompt-based Methods}
Table~\ref{tab:promptresults} shows the results of our proposed method FPT and prompt-based baselines under the few-shot setting. 
(1) Our method FPT 
significantly outperforms nearly all baseline methods across all three datasets under different shots, demonstrating that our method exhibits greater robustness and adaptability to variations in data sizes and languages.
(2) FTP particularly excels on the ChineseLR dataset, and it outperforms the soft prompt (SP) method by 8.41, 14.1, 10.15, 9.94 and 7.9 points under 1, 2, 4, 8, 16 shots, respectively. 
(3) In the task of RA, the soft prompt method generally outperforms the hard prompt. Interestingly, the hybrid prompt, a combination of both, does not always yield better results than the standalone soft prompt. This could be attributed to the inherent challenge in designing and selecting effective hard prompts for RA. Nevertheless, as a hybrid prompt approach that integrates linguistic knowledge, our proposed method continues to exhibit robust performance, demonstrating its adaptability and effectiveness.

\subsection{Comparison with Fusion Methods}
Table~\ref{tab:finetuneresults} reports the experimental results comparing with fusion methods under the few-shot setting. 
(1) Our proposed method FPT shows a stable and significant improvement compared to the 
previous feature fusion methods.  
For instance, in the 2-shot setting, FPT outperforms the best previous fusion methods by 11.28, 5.8 and 11.66 points on ChineseLR, Weebit and Cambridge, respectively. This demonstrates our method's effectiveness in integrating linguistic features for RA.
(2) Methods with linguistic features perform better than standard fine-tuning on Chinese datasets. However, it may not necessarily lead to improvement on English datasets, especially when $k$ is 
increased to a sufficient amount, which indicates
that simply applying linguistic features to aid in English RA is not consistently effective.
\begin{table}[!htbp]
\centering
\small
\resizebox{\linewidth}{!}{
\begin{tabular}{>{\centering\arraybackslash}m{2cm}|>{\raggedright\arraybackslash}m{2.5cm}| ccc}
\toprule
\textbf{Dataset} & \textbf{Methods} & \textbf{k=2} & \textbf{k=4} & \textbf{k=8} \\
\midrule
\multirow{3}{*}{\centering ChineseLR}
& FPT & \textbf{46.24} & \textbf{48.93} & \textbf{52.66} \\
& -SC  & 40.97 & 46.03 & 50.48 \\
& -SC and FP & 25.45 & 36.56 & 40.57 \\
\midrule
\multirow{3}{*}{\centering Weebit}
& FPT & \textbf{55.10} & \textbf{57.70} & \textbf{64.92} \\
& -SC  & 52.68 & 56.92 & 63.63 \\
& -SC and FP & 48.65 & 53.41 & 61.31 \\
\bottomrule
\end{tabular}}
\caption{Ablation study of FPT on ChineseLR and Weebit datasets. SC represents the similarity calibration and FP means utilizing linguistic features as soft prompts.} 
\label{sec:ablation}
\end{table}
\begin{figure}[!htbp]
    \centering
    \includegraphics[width=\columnwidth]{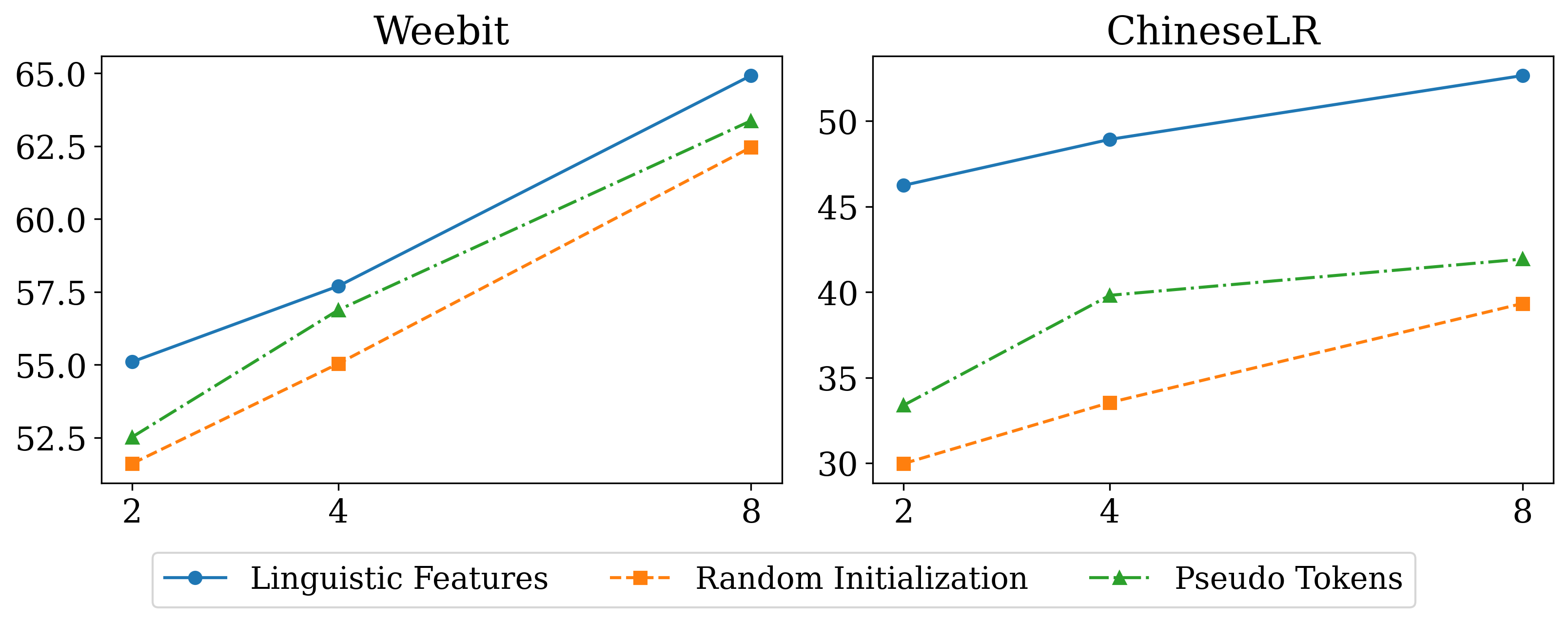}
    \caption{The comparison results of linguistic features, randomly initialized vectors and pseudo tokens.}
    \label{fig:Comparison}
\end{figure}

\subsection{Ablation Study}
To validate the effectiveness of each component in our proposed model, we conduct ablation experiments on both 
English Weebit and ChineseLR datastes. Table~\ref{sec:ablation} lists the results.
Notably, our similarity calibration (SC) is built on the feature prompt (FP), with the aim to maintain consistent inter-class similarity of linguistic features. Therefore, removing FP also detaches SC, explaining why our ablation study is performed incrementally. 

Our full model yields the best performance on both datasets. When removing the SC module, the performance is markedly decreased, demonstrating the necessity of 
retaining the linguistic features' original similarity information during optimization. We have also investigated the impact of SC by visualising the similarity difference matrix before and after applying SC, the results of which are presented in Appendix~\ref{sec:appendiximpactSC}.
Moreover, further removal of the FP shows a steep drop in performance (12.37 points on ChieseLR and 4.29 points on Weebit when $k=4$), validating the effectiveness of incorporating linguistic features as soft prompts.
We note that the improvement of SC and FP is more significant on the Chinese dataset compared to the English dataset, indicating that the Chinese RA task is more dependent on linguistic features.

\subsection{The Significance of Linguistic Features}
To further analyze whether linguistic features improve performance, 
in our model structure, 
we replace the linguistic feature vectors with randomly initialized 
vectors. On the other hand, we re-implement the 
Hybrid Prompt Tuning by utilizing pseudo tokens as soft prompts. We conduct experiments on WeeBit and ChineseLR datasets, and the comparison results are shown in Figure~\ref{fig:Comparison}. 

The performance on both datasets significantly decreases when the linguistic features are replaced with random vectors, especially on the ChineseLR dataset, where the decrease is 
up to 16.27\%. The fewer the samples, the more severe the decline caused by the replacement, further indicating the beneficial role of linguistic features when data is insufficient. Moreover, compared to pseudo tokens, using vector-form embeddings as soft prompts requires the integration of linguistic knowledge to achieve better performance.

\subsection{Comparison with the LLM}
\begin{figure}[!thbp]
    \centering
    \includegraphics[width=\columnwidth]{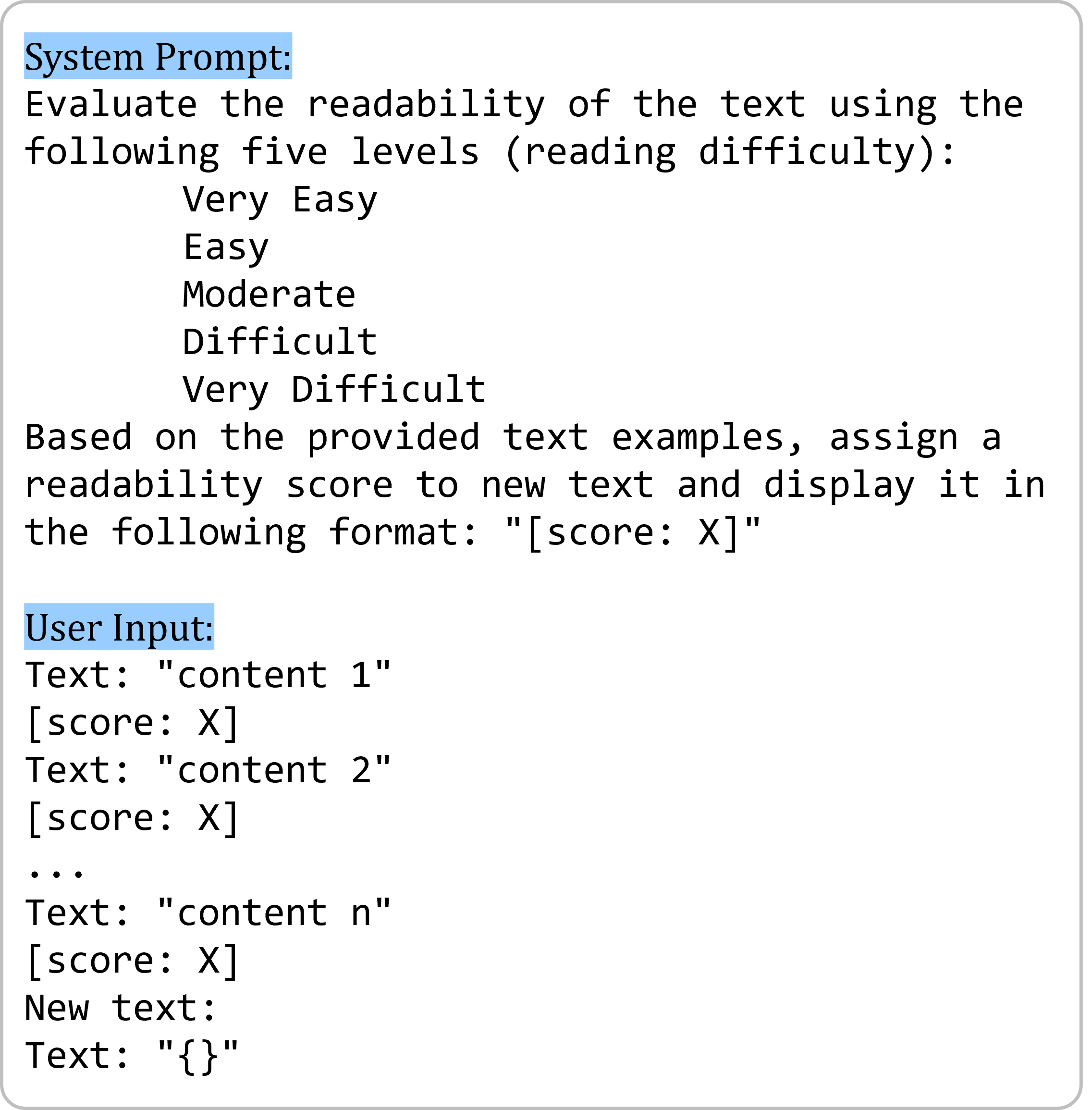}
    \caption{The system prompt and the user input for prompting LLM.}
    \label{fig:llm_prompt}
\end{figure}

The large language model (LLM) 
excels at various downstream tasks without the need for parameter adjustment. We carry out experiments on LLM, utilizing the gpt-3.5-turbo-16k API. We sample the same examples as in other experiments, and the prompt is generated by GPT-4. Specifically, we provide GPT-4 with the task instructions to generate the system prompt and user input for gpt-3.5-turbo-16k, as shown in Figure ~\ref{fig:llm_prompt}. Table~\ref{sec:LLM} shows that our model with 110M parameters 
significantly outperforms the LLM on the English dataset (except for one sample on Cambridge). 
Moreover, gpt-3.5-turbo-16k is unable to perform 1-shot or 2-shot experiments on ChineseLR due to its limited context length. This underscores the necessity for research on handling long texts in RA.

\label{sec:ComLLM}
\begin{table}[!h]
\centering
\small
\begin{tabular}{c|ccc}
\toprule
$k$ & Dataset & FPT  & LLM  \\
\midrule
\multirow{3}{*}{0} & Weebit  & - & 30.79 \\
 & Cambridge & - & 43.33 \\
 & ChineseLR & - & 21.67 \\
\midrule
\multirow{3}{*}{1} & Weebit  & \textbf{43.61} & 31.75 \\
 & Cambridge & 44.17 & \textbf{48.33} \\
 & ChineseLR & 39.63 & -\\
\midrule
\multirow{3}{*}{2} & Weebit  & \textbf{55.10} & 33.17 \\
 & Cambridge & \textbf{59.79} & 54.16 \\
 & ChineseLR & 46.24 & -\\
\bottomrule
\end{tabular}
\caption{Comparison (accuracy) between our model and LLM (gpt-3.5-turbo-16k) on three datasets. $k$ represents the number of 
in-context examples. Due to the limitation of context length, the experiments on Chinese dataset cannot be carried out.}
\label{sec:LLM}
\end{table}




\section{Conclusion}
Inspired by the solid performance of prompt tuning on classification tasks and the importance of linguistic features in the RA task, we empirically investigated the effectiveness of incorporating linguistic features into prompt tuning for RA. We convert linguistic features of the input into soft tokens and utilize the similarity calibration loss to preserve the similarity relationship between classes before and after the transformation. The results show noticeable improvements over previous fusion methods and prompt-based approaches in the few-shot learning setting. The ablation study further illustrated that the proposed model benefits from linguistic features and additional similarity calibration.
Our proposed method, FPT, has demonstrated a new possibility of prompt tuning in an era dominated by LLMs, showcasing its undeniable significance and value in linguistic-related tasks.

\section*{Acknowledgements}
This work is supported by the National Natural Science Foundation of China (62076008) and the Key Project of Natural Science Foundation of China (61936012).

\section*{Limitations}

Our proposed method, which leverages the masked language model (MLM) backbone such as BERT, has demonstrated its efficacy across a variety of natural language processing tasks. Despite its strengths, we acknowledge several limitations that warrant further investigation.

Firstly, our approach exhibits constraints in processing long texts, a scenario frequently encountered in Chinese readability evaluation datasets. The inherent architecture of MLMs like BERT is optimized for shorter sequences, leading to potential performance degradation when dealing with extensive text inputs. 

Secondly, while MLM-based methods are proficient in classification tasks, they often fall short in terms of interpretability of the classification outcomes. The black-box nature of these models makes it challenging to trace and understand the decision-making process, which is crucial for applications where justification of results is required. 

Lastly, the success of our method is significantly contingent upon the quality of linguistic features extracted from the text.  However, the extraction of high-quality linguistic features is not always guaranteed, especially in languages with rich morphology or poor data resources.

In conclusion, while our method stands as a robust approach for several NLP tasks, addressing these limitations is imperative for advancing the field and extending the applicability of MLM-based models to a broader spectrum of text analysis challenges. 
It is also worth noting that only one Chinese dataset is included in this work, as it appears to be the only Chinese RA dataset available to the best of our knowledge. We urge that more attention should be paid to this field of work and further experiments will be conducted if new datasets are released.

\section*{Ethics Statement}

\paragraph{Potential Risks}
Firstly, as a neural network-based method, the predictive outcomes of our approach should not be applied in practical applications without the involvement of human experts. This is a responsible practice for the actual beneficiaries, the learners. Secondly, as mentioned earlier, low-quality or even incorrect linguistic features can negatively impact our method. Therefore, evaluating the quality of linguistic features is essential for the efficacy of our approach.

\paragraph{About Computational Budget}
For each k-shot experiment, we conducted a total of 16 repetitions (refer to Section 5.4) for all baselines and FPT. The duration of a single experiment varies according to the size of k (approximately 20 seconds to 200 seconds), but the time consumed by different methods is almost identical. 

\paragraph{Use of Scientific Artifacts}
We utilize the \textit{lingfeat} toolkit~\cite{lee2021pushing} to extract linguistic features from English texts; this toolkit is publicly accessible under the CC-BY-SA-4.0 license. For extracting Chinese linguistic features, we employ the \textit{zhfeat} toolkit~\cite{li-etal-2022-unified}.

\paragraph{Use of AI Assistants}
We have employed ChatGPT as a writing assistant, primarily for polishing the text after the initial composition.

\bibliography{acl}

\appendix
\section{Details of Linguistic Features}
\label{sec:appendixFeatures}
\subsection{Chinese Linguistic Features}
\begin{center}
\small
\setlength{\tabcolsep}{4pt}
\tablefirsthead{\hline\hline \textbf{Idx} & \textbf{Dim} & \textbf{Feature description} \\ \hline}
\tablehead{\hline\hline \textbf{Idx} & \textbf{Dim} & \textbf{Feature description} \\ \hline}
\tabletail{\hline\hline}
\tablelasttail{\hline\hline}
\bottomcaption{Character features description.}
\begin{supertabular}{p{0.5cm}|m{0.5cm}|p{6cm}}
1 & 1 & Total number of characters \\\hline
2 & 1 & Number of character types \\\hline
3 & 1 & Type Token Ratio (TTR) \\\hline
4 & 1 & Average number of strokes \\\hline
5 & 1 & Weighted average number of strokes \\\hline
6 & 25 & Number of characters with different strokes \\\hline
7 & 25 & Proportion of characters with different strokes \\\hline
8 & 1 & Average character frequency \\\hline
9 & 1 & Weighted average character frequency \\\hline
10 & 1 & Number of single characters \\\hline
11 & 1 & Proportion of single characters \\\hline
12 & 1 & Number of common characters \\\hline
13 & 1 & Proportion of common characters \\\hline
14 & 1 & Number of unregistered characters \\\hline
15 & 1 & Proportion of unregistered characters \\\hline
16 & 1 & Number of first-level characters \\\hline
17 & 1 & Proportion of first-level characters \\\hline
18 & 1 & Number of second-level characters \\\hline
19 & 1 & Proportion of second-level characters \\\hline
20 & 1 & Number of third-level characters \\\hline
21 & 1 & Proportion of third-level characters \\\hline
22 & 1 & Number of fourth-level characters \\\hline
23 & 1 & Proportion of fourth-level characters \\\hline
24 & 1 & Average character level \\
\end{supertabular}
\end{center}

\begin{center}
\small
\setlength{\tabcolsep}{4pt}
\tablefirsthead{\hline\hline \textbf{Idx} & \textbf{Dim} & \textbf{Feature description} \\ \hline}
\tablehead{\hline\hline \textbf{Idx} & \textbf{Dim} & \textbf{Feature description} \\ \hline}
\tabletail{\hline\hline}
\tablelasttail{\hline\hline}
\bottomcaption{Word features description.}
\begin{supertabular}{p{0.5cm}|m{0.5cm}|p{6cm}}
1 & 1 & Total number of words \\\hline
2 & 1 & Number of word types \\\hline
3 & 1 & Type Token Ratio (TTR) \\\hline
4 & 1 & Average word length \\\hline
5 & 1 & Weighted average word length \\\hline
6 & 1 & Average word frequency \\\hline
7 & 1 & Weighted average word frequency \\\hline
8 & 1 & Number of single-character words \\\hline
9 & 1 & Proportion of single-character words \\\hline
10 & 1 & Number of two-character words \\\hline
11 & 1 & Proportion of two-character words \\\hline
12 & 1 & Number of three-character words \\\hline
13 & 1 & Proportion of three-character words \\\hline
14 & 1 & Number of four-character words \\\hline
15 & 1 & Proportion of four-character words \\\hline
16 & 1 & Number of multi-character words \\\hline
17 & 1 & Proportion of multi-character words \\\hline
18 & 1 & Number of idioms \\\hline
19 & 1 & Number of single words \\\hline
20 & 1 & Proportion of single words \\\hline
21 & 1 & Number of unregistered words \\\hline
22 & 1 & Proportion of unregistered words \\\hline
23 & 1 & Number of first-level words \\\hline
24 & 1 & Proportion of first-level words \\\hline
25 & 1 & Number of second-level words \\\hline
26 & 1 & Proportion of second-level words \\\hline
27 & 1 & Number of third-level words \\\hline
28 & 1 & Proportion of third-level words \\\hline
29 & 1 & Number of fourth-level words \\\hline
30 & 1 & Proportion of fourth-level words \\\hline
31 & 1 & Average word level \\\hline
32 & 57 & Number of words with different POS \\\hline
33 & 57 & Proportion of words with different POS \\
\end{supertabular}
\end{center}

\begin{center}
\small
\setlength{\tabcolsep}{4pt}
\tablefirsthead{\hline\hline \textbf{Idx} & \textbf{Dim} & \textbf{Feature description} \\ \hline}
\tablehead{\hline\hline \textbf{Idx} & \textbf{Dim} & \textbf{Feature description} \\ \hline}
\tabletail{\hline\hline}
\tablelasttail{\hline\hline}
\bottomcaption{Sentence features description.}
\begin{supertabular}{p{0.5cm}|m{0.5cm}|p{6cm}}
1 & 1 & Total number of sentences \\\hline
2 & 1 & Average characters in a sentence \\\hline
3 & 1 & Average words in a sentence \\\hline
4 & 1 & Maximum characters in a sentence \\\hline
5 & 1 & Maximum words in a sentence \\\hline
6 & 1 & Number of clauses \\\hline
7 & 1 & Average characters in a clause \\\hline
8 & 1 & Average words in a clause \\\hline
9 & 1 & Maximum characters in a clause \\\hline
10 & 1 & Maximum words in a clause \\\hline
11 & 30 & Sentence length distribution \\\hline
12 & 1 & Average syntax tree height \\\hline
13 & 1 & Maximum syntax tree height \\\hline
14 & 1 & Syntax tree height <= 5 ratio \\\hline
15 & 1 & Syntax tree height <= 10 ratio \\\hline
16 & 1 & Syntax tree height <= 15 ratio \\\hline
17 & 1 & Syntax tree height >= 16 ratio \\\hline
18 & 14 & Dependency distribution  \\\hline
\end{supertabular}
\end{center}

\begin{center}
\small
\setlength{\tabcolsep}{4pt}
\tablefirsthead{\hline\hline \textbf{Idx} & \textbf{Dim} & \textbf{Feature description} \\ \hline}
\tablehead{\hline\hline \textbf{Idx} & \textbf{Dim} & \textbf{Feature description} \\ \hline}
\tabletail{\hline\hline}
\tablelasttail{\hline\hline}
\bottomcaption{Paragraph features description.}
\begin{supertabular}{p{0.5cm}|m{0.5cm}|p{6cm}}
1 & 1 & Total number of paragraphs \\\hline
2 & 1 & Average characters in a paragraph \\\hline
3 & 1 & Average words in a paragraph \\\hline
4 & 1 & Maximum characters in a paragraph \\\hline
5 & 1 & Maximum words in a paragraph \\
\end{supertabular}
\end{center}

\subsection{English Linguistic Features}
\begin{center}
\small
\setlength{\tabcolsep}{4pt}
\tablefirsthead{\hline\hline \textbf{Idx} & \textbf{Dim} & \textbf{Feature description} \\ \hline}
\tablehead{\hline\hline \textbf{Idx} & \textbf{Dim} & \textbf{Feature description} \\ \hline}
\tabletail{\hline\hline}
\tablelasttail{\hline\hline}
\bottomcaption{Discourse features description.}
\begin{supertabular}{p{0.5cm}|m{0.5cm}|p{6cm}}
1 & 1 & Total number of Entities Mentions counts \\\hline
2 & 1 & Average number of Entities Mentions counts per sentence \\\hline
3 & 1 & Average number of Entities Mentions counts per token (word) \\\hline
4 & 1 & Total number of unique Entities \\\hline
5 & 1 & Average number of unique Entities per sentence \\\hline
6 & 1 & Average number of Entities Mentions counts per token (word)s \\\hline
7 & 1 & Total number of unique Entities \\\hline
8 & 1 & Ratio of ss transitions to total \\\hline
9 & 1 & Ratio of so transitions to total \\\hline
10 & 1 & Ratio of sx transitions to total \\\hline
11 & 1 & Ratio of sn transitions to total \\\hline
12 & 1 & Ratio of os transitions to total \\\hline
13 & 1 & Ratio of oo transitions to total \\\hline
14 & 1 & Ratio of ox transitions to total \\\hline
15 & 1 & Ratio of on transitions to total \\\hline
16 & 1 & Ratio of xs transitions to total \\\hline
17 & 1 & Ratio of xo transitions to total \\\hline
18 & 1 & Ratio of xx transitions to total \\\hline
19 & 1 & Ratio of xn transitions to total \\\hline
20 & 1 & Ratio of ns transitions to total \\\hline
21 & 1 & Ratio of no transitions to total \\\hline
22 & 1 & Ratio of nx transitions to total \\\hline
23 & 1 & Ratio of nn transitions to total \\\hline
24 & 1 & Local Coherence for PA score \\\hline
25 & 1 & Local Coherence for PW score \\\hline
26 & 1 & Local Coherence for PU score \\\hline
27 & 1 & Local Coherence distance for PA score \\\hline
28 & 1 & Local Coherence distance for PW score \\\hline
29 & 1 & Local Coherence distance for PU score \\\hline
\end{supertabular}
\end{center}

\begin{center}
\small
\setlength{\tabcolsep}{4pt}
\tablefirsthead{\hline\hline \textbf{Idx} & \textbf{Dim} & \textbf{Feature description} \\ \hline}
\tablehead{\hline\hline \textbf{Idx} & \textbf{Dim} & \textbf{Feature description} \\ \hline}
\tabletail{\hline\hline}
\tablelasttail{\hline\hline}
\bottomcaption{Syntactic features description.}
\begin{supertabular}{p{0.5cm}|m{0.5cm}|p{6cm}}
1 & 1 & Total count of Noun phrases \\\hline
2 & 1 & Average count of Noun phrases per sentence \\\hline
3 & 1 & Average count of Noun phrases per token \\\hline
4 & 1 & Ratio of Noun phrases count to Verb phrases count \\\hline
5 & 1 & Ratio of Noun phrases count to Subordinate Clauses count \\\hline
6 & 1 & Ratio of Noun phrases count to Prep phrases count \\\hline
7 & 1 & Ratio of Noun phrases count to Adj phrases count \\\hline
8 & 1 & Ratio of Noun phrases count to Adv phrases count \\\hline
9 & 1 & Total count of Verb phrases \\\hline
10 & 1 & Average count of Verb phrases per sentence \\\hline
11 & 1 & Average count of Verb phrases per token \\\hline
12 & 1 & Ratio of Verb phrases count to Noun phrases count \\\hline
13 & 1 & Ratio of Verb phrases count to Subordinate Clauses count \\\hline
14 & 1 & Ratio of Verb phrases count to Prep phrases count \\\hline
15 & 1 & Ratio of Verb phrases count to Adj phrases count \\\hline
16 & 1 & Ratio of Verb phrases count to Adv phrases count \\\hline
17 & 1 & Total count of Subordinate Clauses \\\hline
18 & 1 & Average count of Subordinate Clauses per sentence \\\hline
19 & 1 & Average count of Subordinate Clauses per token \\\hline
20 & 1 & Ratio of Subordinate Clauses count to Noun phrases count \\\hline
21 & 1 & Ratio of Subordinate Clauses count to Verb phrases count \\\hline
22 & 1 & Ratio of Subordinate Clauses count to Prep phrases count \\\hline
23 & 1 & Ratio of Subordinate Clauses count to Adj phrases count \\\hline
24 & 1 & Ratio of Subordinate Clauses count to Adv phrases count \\\hline
25 & 1 & Total count of prepositional phrases\\\hline
26 & 1 & Average count of prepositional phrases per sentence \\\hline
27 & 1 & Average count of prepositional phrases per token \\\hline
28 & 1 & Ratio of Prep phrases count to Noun phrases count \\\hline
29 & 1 & Ratio of Prep phrases count to Verb phrases count \\\hline
30 & 1 & Ratio of Prep phrases count to Subordinate Clauses count \\\hline
31 & 1 & Ratio of Prep phrases count to Adj phrases count \\\hline
32 & 1 & Ratio of Prep phrases count to Adv phrases count \\\hline
33 & 1 & Total count of Adjective phrases \\\hline
34 & 1 & Average count of Adjective phrases per sentence \\\hline
35 & 1 & Average count of Adjective phrases per token \\\hline
36 & 1 & Ratio of Adj phrases count to Noun phrases count \\\hline
37 & 1 & Ratio of Adj phrases count to Verb phrases count \\\hline
38 & 1 & Ratio of Adj phrases count to Subordinate Clauses count \\\hline
39 & 1 & Ratio of Adj phrases count to Prep phrases count \\\hline
40 & 1 & Ratio of Adj phrases count to Adv phrases count \\\hline
41 & 1 & Total count of Adverb phrases \\\hline
42 & 1 & Average count of Adverb phrases per sentence \\\hline
43 & 1 & Average count of Adverb phrases per token \\\hline
44 & 1 & Ratio of Adv phrases count to Noun phrases count \\\hline
45 & 1 & Ratio of Adv phrases count to Verb phrases count \\\hline
46 & 1 & Ratio of Adv phrases count to Subordinate Clauses count \\\hline
47 & 1 & Ratio of Adv phrases count to Prep phrases count \\\hline
48 & 1 & Ratio of Adv phrases count to Adj phrases count \\\hline
49 & 1 & Total Tree height of all sentences \\\hline
50 & 1 & Average Tree height per sentence\\\hline
51 & 1 & Average Tree height per token (word) \\\hline
52 & 1 & Total length of flattened Trees \\\hline
53 & 1 & Average length of flattened Trees per sentence\\\hline
54 & 1 & Average length of flattened Trees per token (word)\\\hline
55 & 1 & Total count of Noun POS tags\\\hline
56 & 1 & Average count of Noun POS tags per sentence\\\hline
57 & 1 & Average count of Noun POS tags per token\\\hline
58 & 1 & Ratio of Noun POS count to Adjective POS count \\\hline
59 & 1 & Ratio of Noun POS count to Verb POS count \\\hline
60 & 1 & Ratio of Noun POS count to Adverb POS count\\\hline
61 & 1 & Ratio of Noun POS count to Subordinating Conjunction count \\\hline
62 & 1 & Ratio of Noun POS count to Coordinating Conjunction count \\\hline
63 & 1 & Total count of Verb POS tags \\\hline
64 & 1 & Average count of Verb POS tags per sentence \\\hline
65 & 1 & Average count of Verb POS tags per token \\\hline
66 & 1 & Ratio of Verb POS count to Adjective POS count \\\hline
67 & 1 & Ratio of Verb POS count to Noun POS count \\\hline
68 & 1 & Ratio of Verb POS count to Adverb POS count \\\hline
69 & 1 & Ratio of Verb POS count to Subordinating Conjunction count \\\hline
70 & 1 & Ratio of Verb POS count to Coordinating Conjunction count \\\hline
71 & 1 & Total count of Adjective POS tags \\\hline
72 & 1 & Average count of Adjective POS tags per sentence \\\hline
73 & 1 & Average count of Adjective POS tags per token \\\hline
74 & 1 & Ratio of Adjective POS count to Noun POS count \\\hline
75 & 1 & Ratio of Adjective POS count to Verb POS count \\\hline
76 & 1 & Ratio of Adjective POS count to Adverb POS count \\\hline
77 & 1 & Ratio of Adjective POS count to Subordinating Conjunction count \\\hline
78 & 1 & Ratio of Adjective POS count to Coordinating Conjunction count \\\hline
79 & 1 & Total count of Adverb POS tags \\\hline
80 & 1 & Average count of Adverb POS tags per sentence \\\hline
81 & 1 & Average count of Adverb POS tags per token \\\hline
82 & 1 & Ratio of Adverb POS count to Adjective POS count \\\hline
83 & 1 & Ratio of Adverb POS count to Noun POS count \\\hline
84 & 1 & Ratio of Adverb POS count to Verb POS count \\\hline
85 & 1 & Ratio of Adverb POS count to Subordinating Conjunction count \\\hline
86 & 1 & Ratio of Adverb POS count to Coordinating Conjunction count \\\hline
87 & 1 & Total count of Subordinating Conjunction POS tags \\\hline
88 & 1 & Average count of Subordinating Conjunction POS tags per sentence \\\hline
89 & 1 & Average count of Subordinating Conjunction POS tags per token \\\hline
90 & 1 & Ratio of Subordinating Conjunction POS count to Adjective POS count \\\hline
91 & 1 & Ratio of Subordinating Conjunction POS count to Noun POS count \\\hline
92 & 1 & Ratio of Subordinating Conjunction POS count to Verb POS count \\\hline
93 & 1 & Ratio of Subordinating Conjunction POS count to Adverb POS count \\\hline
94 & 1 & Ratio of Subordinating Conjunction POS count to Coordinating Conjunction count \\\hline
95 & 1 & Total count of Coordinating Conjunction POS tags \\\hline
96 & 1 & Average count of Coordinating Conjunction POS tags per sentence \\\hline
97 & 1 & Average count of Coordinating Conjunction POS tags per token \\\hline
98 & 1 & Ratio of Coordinating Conjunction POS count to Adjective POS count \\\hline
99 & 1 & Ratio of Coordinating Conjunction POS count to Noun POS count \\\hline
100 & 1 & Ratio of Coordinating Conjunction POS count to Verb POS count \\\hline
101 & 1 & Ratio of Coordinating Conjunction POS count to Adverb POS count \\\hline
102 & 1 & Ratio of Coordinating Conjunction POS count to Subordinating Conjunction count \\\hline
103 & 1 & Total count of Content words \\\hline
104 & 1 & Average count of Content words per sentence \\\hline
105 & 1 & Average count of Content words per token \\\hline
106 & 1 & Total count of Function words \\\hline
107 & 1 & Average count of Function words per sentence \\\hline
108 & 1 & Average count of Function words per token \\\hline
109 & 1 & Ratio of Content words to Function words \\\hline
\end{supertabular}
\end{center}

\begin{center}
\small
\setlength{\tabcolsep}{4pt}
\tablefirsthead{\hline\hline \textbf{Idx} & \textbf{Dim} & \textbf{Feature description} \\ \hline}
\tablehead{\hline\hline \textbf{Idx} & \textbf{Dim} & \textbf{Feature description} \\ \hline}
\tabletail{\hline\hline}
\tablelasttail{\hline\hline}
\bottomcaption{Lexico Semantic features description.}
\begin{supertabular}{p{0.5cm}|m{0.5cm}|p{6cm}}
1 & 1 & Unique Nouns/total Nouns (Noun Variation-1) \\\hline
2 & 1 & (Unique Nouns**2)/total Nouns (Squared Noun Variation-1) \\\hline
3 & 1 & Unique Nouns/sqrt(2*total Nouns) (Corrected Noun Variation-1) \\\hline
4 & 1 & Unique Verbs/total Verbs (Verb Variation-1) \\\hline
5 & 1 & (Unique Verbs**2)/total Verbs (Squared Verb Variation-1) \\\hline
6 & 1 & Unique Verbs/sqrt(2*total Verbs) (Corrected Verb Variation-1) \\\hline
7 & 1 & Unique Adjectives/total Adjectives (Adjective Variation-1) \\\hline
8 & 1 & (Unique Adjectives**2)/total Adjectives (Squared Adjective Variation-1) \\\hline
9 & 1 & Unique Adjectives/sqrt(2*total Adjectives) (Corrected Adjective Variation-1) \\\hline
10 & 1 & Unique Adverbs/total Adverbs (AdVerb Variation-1) \\\hline
11 & 1 & (Unique Adverbs**2)/total Adverbs (Squared AdVerb Variation-1) \\\hline
12 & 1 & Unique Adverbs/sqrt(2*total Adverbs) (Corrected AdVerb Variation-1) \\\hline
13 & 1 & Unique tokens/total tokens (TTR) \\\hline
14 & 1 & Unique tokens/sqrt(2*total tokens) (Corrected TTR) \\\hline
15 & 1 & Log(unique tokens)/log(total tokens) (Bi-Logarithmic TTR)
 \\\hline
16 & 1 & (Log(unique tokens))**2/log(total tokens/unique tokens) (Uber Index) \\\hline
17 & 1 & Measure of Textual Lexical Diversity (default TTR = 0.72) \\\hline
18 & 1 & Total AoA (Age of Acquisition) of words \\\hline
19 & 1 & Average AoA of words per sentence \\\hline
20 & 1 & Average AoA of words per token \\\hline
21 & 1 & Total lemmas AoA of lemmas \\\hline
22 & 1 & Average lemmas AoA of lemmas per sentence \\\hline
23 & 1 & Average lemmas AoA of lemmas per token \\\hline
24 & 1 & Total lemmas AoA of lemmas, Bird norm \\\hline
25 & 1 & Average lemmas AoA of lemmas, Bird norm per sentence \\\hline
26 & 1 & Average lemmas AoA of lemmas, Bird norm per token \\\hline
27 & 1 & Total lemmas AoA of lemmas, Bristol norm \\\hline
28 & 1 & Average lemmas AoA of lemmas, Bristol norm per sentence \\\hline
29 & 1 & Average lemmas AoA of lemmas, Bristol norm per token \\\hline
30 & 1 & Total AoA of lemmas, Cortese and Khanna norm \\\hline
31 & 1 & Average AoA of lemmas, Cortese and Khanna norm per sentence \\\hline
32 & 1 & Average AoA of lemmas, Cortese and Khanna norm per token \\\hline
33 & 1 & Total SubtlexUS FREQcount value \\\hline
34 & 1 & Average SubtlexUS FREQcount value per sentenc \\\hline
35 & 1 & Average SubtlexUS FREQcount value per token \\\hline
36 & 1 & Total SubtlexUS CDcount value \\\hline
37 & 1 & Average SubtlexUS CDcount value per sentence \\\hline
38 & 1 & Average SubtlexUS CDcount value per token \\\hline
39 & 1 & Total SubtlexUS FREQlow value \\\hline
40 & 1 & Average SubtlexUS FREQlow value per sentence \\\hline
41 & 1 & Average SubtlexUS FREQlow value per token \\\hline
42 & 1 & Total SubtlexUS CDlow value \\\hline
43 & 1 & Average SubtlexUS CDlow value per sentence \\\hline
44 & 1 & Average SubtlexUS CDlow value per token \\\hline
45 & 1 & Total SubtlexUS SUBTLWF value \\\hline
46 & 1 & Average SubtlexUS SUBTLWF value per sentence \\\hline
47 & 1 & Average SubtlexUS SUBTLWF value per token \\\hline
48 & 1 & Total SubtlexUS Lg10WF value \\\hline
49 & 1 & Average SubtlexUS Lg10WF value per sentence \\\hline
50 & 1 & Average SubtlexUS Lg10WF value per token \\\hline
51 & 1 & Total SubtlexUS SUBTLCD value \\\hline
52 & 1 & Average SubtlexUS SUBTLCD value per sentence \\\hline
53 & 1 & Average SubtlexUS SUBTLCD value per token \\\hline
54 & 1 & Total SubtlexUS Lg10CD value \\\hline
55 & 1 & Average SubtlexUS Lg10CD value per sentence \\\hline
56 & 1 & Average SubtlexUS Lg10CD value per token \\\hline
\end{supertabular}
\end{center}

\begin{center}
\small
\setlength{\tabcolsep}{4pt}
\tablefirsthead{\hline\hline \textbf{Idx} & \textbf{Dim} & \textbf{Feature description} \\ \hline}
\tablehead{\hline\hline \textbf{Idx} & \textbf{Dim} & \textbf{Feature description} \\ \hline}
\tabletail{\hline\hline}
\tablelasttail{\hline\hline}
\bottomcaption{Shallow Traditional features description.}
\begin{supertabular}{p{0.5cm}|m{0.5cm}|p{6cm}}
1 & 1 & Total count of tokens x total count of sentence \\\hline
2 & 1 & Sqrt(total count of tokens x total count of sentence) \\\hline
3 & 1 & Log(total count of tokens)/log(total count of sentence) \\\hline
4 & 1 & Average count of tokens per sentence \\\hline
5 & 1 & Average count of syllables per sentence \\\hline
6 & 1 & Average count of syllables per token \\\hline
7 & 1 & Average count of characters per sentence \\\hline
8 & 1 & Average count of characters per token \\\hline
9 & 1 & Smog Index \\\hline
10 & 1 & Coleman Liau Readability Score \\\hline
11 & 1 & Gunning Fog Count Score \\\hline
12 & 1 & New Automated Readability Index \\\hline
13 & 1 & Flesch Kincaid Grade Level \\\hline
14 & 1 & Linsear Write Formula Score \\\hline
\end{supertabular}
\end{center}

\section{Templates}
\label{sec:appendixTemplates}
\textbf{Chinese Dataset } Based on the \textit{Chinese Curriculum Standards for Compulsory Education}, we devise the following templates:
\begin{CJK*}{UTF8}{gbsn}
\begin{itemize}

    \item $T_1(\cdot)$ = 一篇第[MASK]学段的文章: {}
    \item $T_2(\cdot)$ = 这是一篇第[MASK]学段的课文: {}
    \item $T_3(\cdot)$ = 一篇第[MASK]学段的课文: {}
    \item $T_4(\cdot)$ = 一篇阅读难度为[MASK]的课文: {}

\end{itemize}
\end{CJK*}
\textbf{English Dataset } Based on ~\cite{vajjala2012improving}, we use the following templates:
\begin{itemize}
    \item $T_1(\cdot)$ = A [MASK] article to understand: {}
    \item $T_2(\cdot)$ = A [MASK] text to understand: {}
    \item $T_3(\cdot)$ = This is a [MASK] article to understand: {}
    \item $T_4(\cdot)$ = A [MASK] article to read: {}
\end{itemize}

\section{The Impact of Similarity Calibration}
\label{sec:appendiximpactSC}
\begin{figure}[!h]
  \centering
  \subfigure[ChineseLR w/o SC]{
    \includegraphics[width=0.17\textwidth]{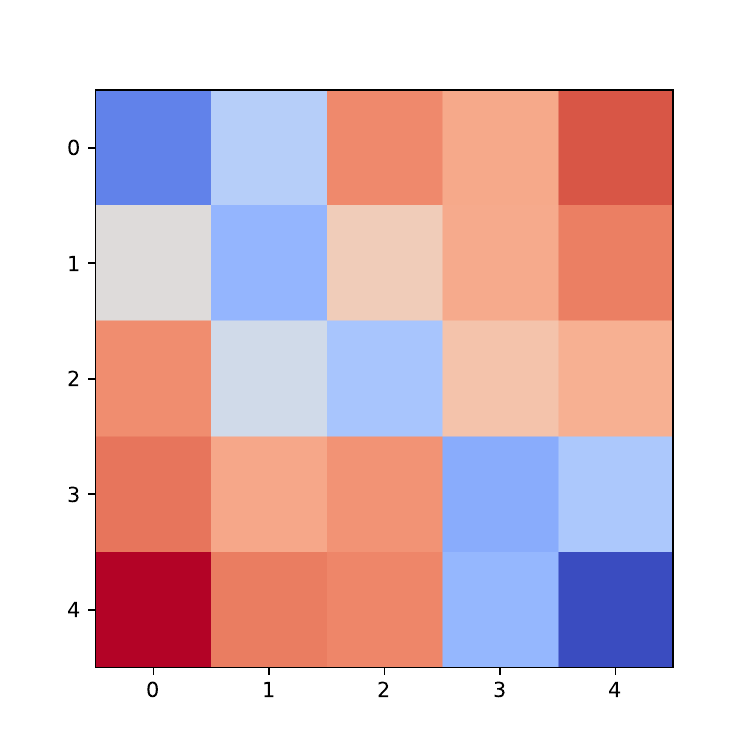}
  }
  \subfigure[ChineseLR w/ SC]{
    \includegraphics[width=0.17\textwidth]{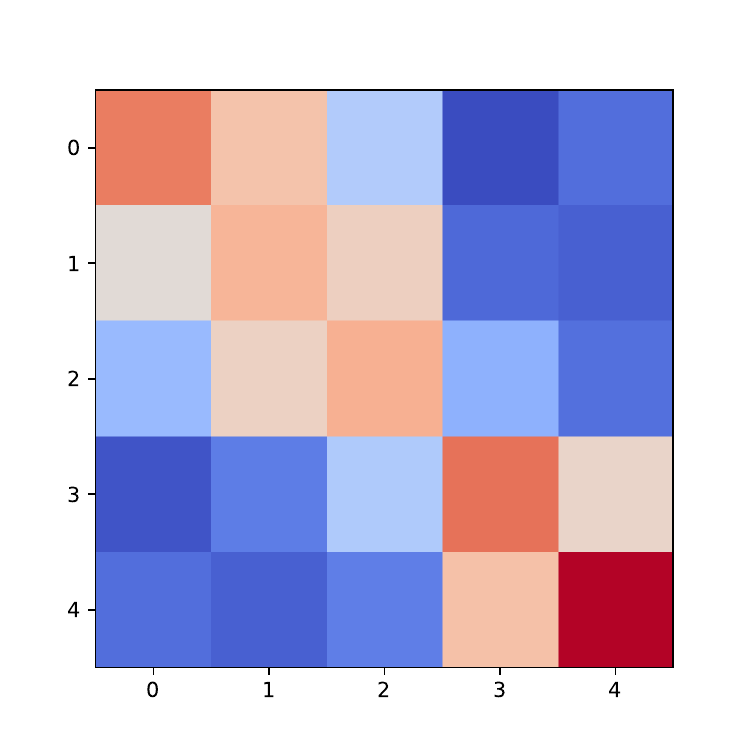}
  }
  \includegraphics[width=0.10\textwidth]{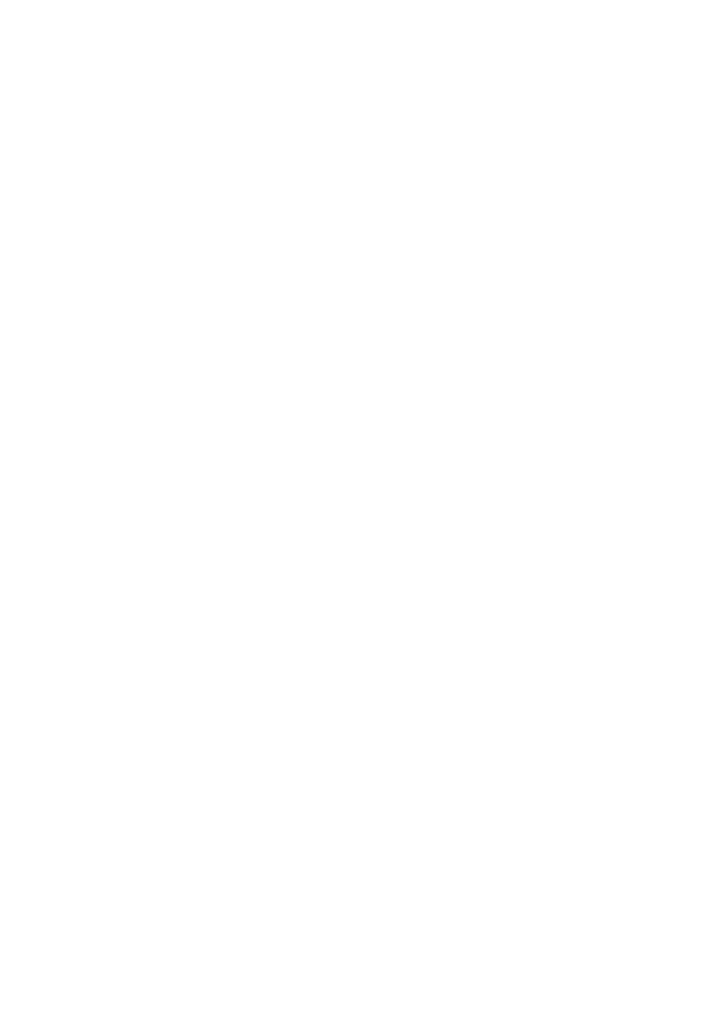}
  \subfigure[Weebit w/o SC]{
    \includegraphics[width=0.17\textwidth]{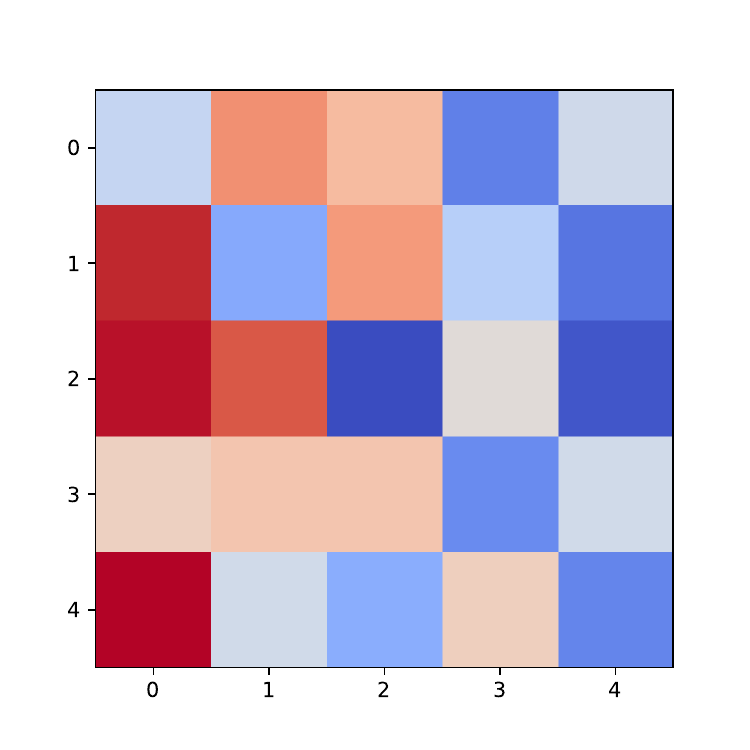}
  }
  \subfigure[Weebit w/ SC]{
    \includegraphics[width=0.17\textwidth]{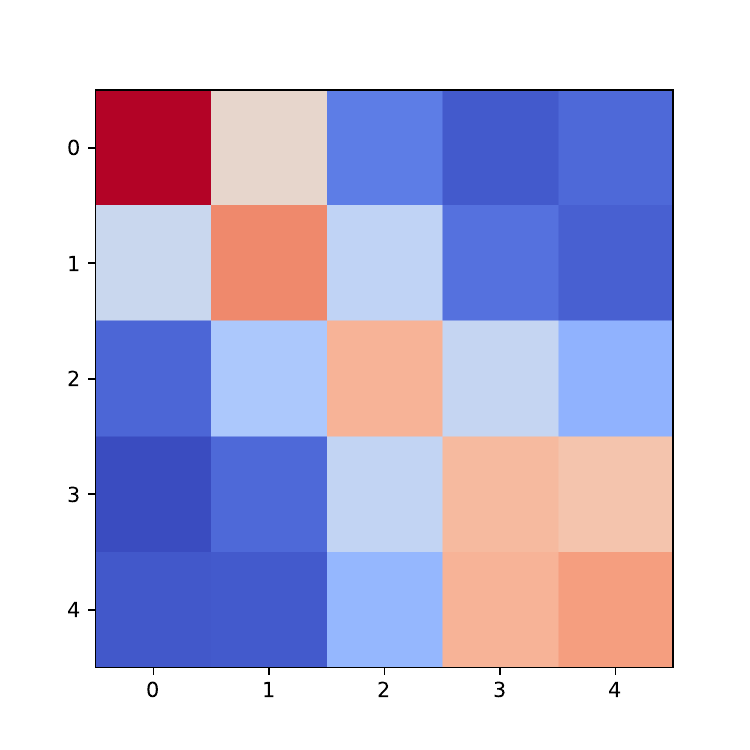}
  }
  \includegraphics[width=0.10\textwidth]{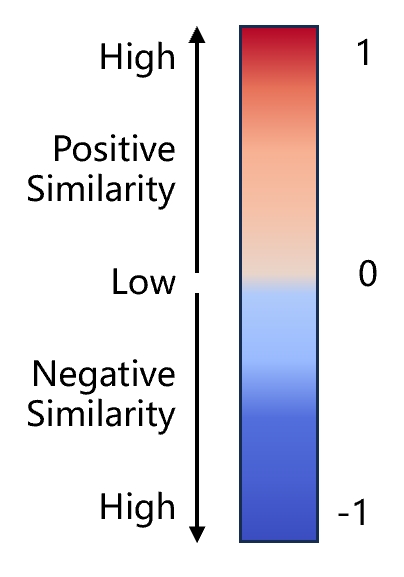}
  \caption{
  Similarity difference matrices. We plot the difference matrices of similarity before and after linguistic feature embedding, both with and without SC. The horizontal and vertical coordinates represent the level of linguistic features. By comparing the diagonal of the matrix before and after the similarity calibration (that is, the similarity between linguistic features of the same level), the similarity between analogous categories is drawn closer.}
  \label{sc}
\end{figure}

To investigate the impact of Similarity Calibration (SC), we plot the similarity difference matrices before and after linguistic feature embedding on two datasets, both with and without SC. 
Specifically, we calculate the similarity of linguistic features between each category before and after embedding to obtain two similarity matrices.
Then we subtract the former from the latter to obtain the difference matrix. The results are shown in Figure~\ref{sc}, 
where the diagonal of the matrix represents the similarity of the linguistic features from the same category. 

On both datasets, SC can effectively increase the similarity between the same and analogous categories (represented by warm colors), while reducing the similarity between distance categories (represented by cool colors). 
This can provide effective assistance for classification tasks.


% ... 

\end{document}